\documentclass[conference]{IEEEtran}
\usepackage{amsmath}

\usepackage{cite}
\usepackage{amsmath,amssymb,amsfonts}
\usepackage{algorithmic}
\usepackage{graphicx}
\usepackage{textcomp}
\usepackage{xcolor}
\usepackage[T1]{fontenc}
\usepackage[utf8]{inputenc}
\usepackage{float}
\def\BibTeX{{\rm B\kern-.05em{\sc i\kern-.025em b}\kern-.08em
    T\kern-.1667em\lower.7ex\hbox{E}\kern-.125emX}}

\usepackage{fancyhdr}

\begin{document}

\title{Accurate Localization of Road Traffic Objects on the Road Plane Using Surveillance Camera Imagery}

\author{\IEEEauthorblockN{1\textsuperscript{st} Jan Gawroński, BSc}
\IEEEauthorblockA{\textit{Warsaw University of Technology} \\
\textit{Faculty of Electrical Engineering}\\
Warsaw, Poland\\
jan.gawronski.stud@pw.edu.pl
}
\and
\IEEEauthorblockN{2\textsuperscript{nd} Witold Czajewski, PhD }
\IEEEauthorblockA{\textit{Warsaw University of Technology} \\
\textit{Faculty of Electrical Engineering}\\
Warsaw, Poland\\
witold.czajewski@pw.edu.pl}

}

\maketitle
\thispagestyle{fancy}

\begin{abstract}

Accurate vehicle localization from monocular roadside surveillance cameras is an important problem in intelligent transportation systems, traffic monitoring, and traffic conflict analysis. Standard localization approaches typically estimate vehicle position using the center of the detector bounding box, which may lead to large localization errors due to perspective distortion and parallax effects, particularly for elevated roadside cameras and large vehicles.

This paper proposes a two-stage geometry-aware localization pipeline that estimates the projection of the vehicle footprint onto the road plane instead of relying directly on detector geometry. In the first stage, vehicles are detected using a YOLO26-based detector. In the second stage, a dedicated ResNet34 regression network predicts four corner points corresponding to the projection of the vehicle base onto the image plane. The final vehicle position is estimated as the geometric center of the predicted quadrilateral.

The proposed method was trained using synthetic data generated in the CARLA simulation environment and subsequently fine-tuned on real-world roadside imagery from the DAIR-V2X dataset. Experimental evaluation performed on both synthetic and real-world data demonstrated clear improvements in localization accuracy compared with naive bounding-box-center-based localization. On the DAIR-V2X dataset, the proposed approach reduced the mean image-space localization error from 31.77~px to 15.30~px (51.8\% improvement) and the median error to 4.29~px. Median ground-plane localization error for medium-range vehicles decreased from 5.52~m to 0.90~m, while for far-range vehicles it decreased from 8.67~m to 1.84~m.

The experiments additionally demonstrated that contextual information surrounding the detector bounding box plays an important role in geometric localization. The largest improvements were observed for distant vehicles and geometrically challenging cases affected by strong perspective distortion and parallax effects.

\end{abstract}


\section{Introduction}

Accurate localization of road traffic participants from monocular surveillance cameras is an important problem in intelligent transportation systems (ITS), traffic monitoring, and urban infrastructure analytics \cite{xinyao2023centerloc3dmonocular3dvehicle,dair}. In many practical applications, such as traffic flow estimation, trajectory reconstruction, near-collision analysis, adaptive traffic signal control, and speed estimation, it is necessary not only to detect vehicles in an image, but also to determine their position on the road plane with high precision. Accurate ground-plane localization is particularly important in traffic conflict analysis and surrogate safety assessment, where small localization errors may significantly affect estimated trajectories, conflict points, or time-to-collision metrics \cite{dabkowski_video,dabkowski_pdi}.

Compared with vehicle-mounted sensing systems, roadside monocular cameras offer several practical advantages, including low deployment cost, large field of view, and compatibility with existing urban monitoring infrastructure \cite{dair}. As a result, roadside vision systems have become an increasingly important research direction in recent years \cite{kim2019deep,xinyao2023centerloc3dmonocular3dvehicle}. However, precise localization from monocular roadside imagery remains challenging due to perspective distortion, occlusion, truncation, varying object scales, and strong parallax effects caused by elevated camera placement \cite{kim2019deep,kim2023cnn,xinyao2023centerloc3dmonocular3dvehicle}.

In standard object detection pipelines, vehicle position is commonly approximated using either the center of the detector bounding box or the midpoint of its bottom edge \cite{electronics14071291,kim2023cnn}. While computationally simple, these representations do not correspond to the true projection of the vehicle onto the road plane. The discrepancy becomes particularly severe for elevated roadside cameras observing tall vehicles or scenes captured at large viewing angles relative to the road surface. As a consequence, even relatively small localization errors in image space may translate into several metres of error after projection onto the road plane.

To address this problem, this paper proposes a two-stage localization pipeline that estimates the projection of the vehicle footprint onto the road surface instead of relying directly on detector geometry. In the first stage, vehicles are detected using a YOLO26-based detector \cite{yolo26_ultralytics}. In the second stage, a dedicated regression network predicts four characteristic points corresponding to the projection of the vehicle base onto the road plane. These points define a quadrilateral approximating the contact region between the vehicle and the road surface. The final vehicle position is estimated as the geometric center of the predicted quadrilateral, providing a representation that is more consistent with the true ground-plane location of the vehicle and reducing localization error caused by perspective distortion and parallax.

The proposed approach separates detection and geometric regression into two independent stages. This allows the regression network to focus exclusively on geometric estimation while also enabling the use of enlarged image crops containing contextual information outside the detector bounding box. The method is trained using synthetic data generated in the CARLA simulation environment \cite{dosovitskiy2017carla} and subsequently fine-tuned on real-world roadside imagery from the DAIR-V2X dataset \cite{dair}.

The proposed method is evaluated on both synthetic and real-world datasets using pixel-domain and ground-plane localization metrics. Experimental results demonstrate substantial reductions in localization error compared with standard bounding-box-based approaches. In particular, the proposed method significantly improves localization accuracy for distant vehicles and for large vehicles affected by strong parallax distortion.

The main contributions of this work are as follows:

\begin{itemize}
    \item A two-stage monocular vehicle localization pipeline for roadside surveillance cameras combining object detection and geometric footprint regression.
    
    \item A regression-based representation of the projected vehicle base quadrilateral for improved road-plane localization.
    
    \item An analysis of the influence of contextual crop enlargement on regression quality.
    
    \item A comprehensive evaluation on both synthetic CARLA data and real-world DAIR-V2X imagery using image-space and metric road-plane localization errors.
    
    \item An experimental analysis of localization accuracy as a function of vehicle distance, visibility conditions, and vehicle category.
\end{itemize}

\section{Related Work}

Vehicle localization from roadside surveillance cameras is commonly performed using standard 2D object detectors such as YOLO or Faster R-CNN, where the object position is approximated using either the center of the detector bounding box or the midpoint of its bottom edge. While computationally efficient, these representations often introduce considerable localization errors because they do not correspond to the true projection of the vehicle onto the road plane \cite{electronics14071291}. The problem becomes particularly pronounced for elevated roadside cameras, large viewing angles, and tall vehicles, where perspective distortion and parallax effects are strong.

One common approach for improving localization accuracy is the use of geometric transformations such as inverse perspective mapping (IPM) or homography-based projection methods \cite{kim2019deep,zhumonocular}. These methods transform the image into a bird’s-eye-view representation, reducing perspective distortion and enabling more accurate estimation of vehicle position and orientation. However, such approaches typically require accurate camera calibration, manually selected landmarks, or assumptions about the road geometry. In practice, their performance may degrade when calibration is inaccurate or when the road surface deviates from a planar model.

More advanced approaches attempt to estimate full 3D object properties directly from monocular imagery. CenterFusion \cite{nabati2021centerfusion}, for example, combines radar and camera data for 3D object detection, improving localization robustness at the cost of requiring additional sensing hardware. Another related approach is CenterLoc3D \cite{xinyao2023centerloc3dmonocular3dvehicle}, which predicts the vehicle centroid and 3D bounding box vertices in image space and subsequently reconstructs object position in world coordinates. Such methods provide richer geometric information, but are typically more computationally demanding and involve significantly more complex prediction targets.

The work most closely related to the proposed approach is the BFQ-based method introduced by Kim et al. \cite{kim2023cnn}, where additional regression heads are integrated into a YOLOv4 detector to estimate the vehicle bottom face quadrilateral. Similarly, Kim et al. \cite{electronics14071291} proposed extracting object-specific key points corresponding to ground-contact locations for improved road-user localization. These approaches demonstrate that explicitly modeling object geometry can significantly improve localization accuracy compared with standard bounding-box-based representations.

The proposed method shares the general idea of geometry-aware localization from monocular roadside imagery, but differs in several important aspects. Instead of integrating geometric regression directly into the detector, the proposed approach separates detection and geometric estimation into two independent stages. A dedicated ResNet34-based regression network is applied to enlarged image crops generated from YOLO26 detections, allowing the regression stage to focus exclusively on geometric estimation while also exploiting contextual information outside the detector bounding box. In contrast to full monocular 3D reconstruction methods, the proposed approach estimates only the projection of the vehicle footprint onto the road plane, which simplifies the prediction problem while still providing accurate ground-plane localization.

Furthermore, direct quantitative comparison with several existing approaches remains difficult due to limited reproducibility and the absence of publicly available implementations fully matching the published experimental setups. Therefore, the evaluation in this work focuses primarily on comparison against standard bounding-box-based localization approaches commonly used in practical surveillance systems.

\section{Proposed Method}

\begin{figure*}[t]
    \centering
    \includegraphics[width=.9\textwidth]{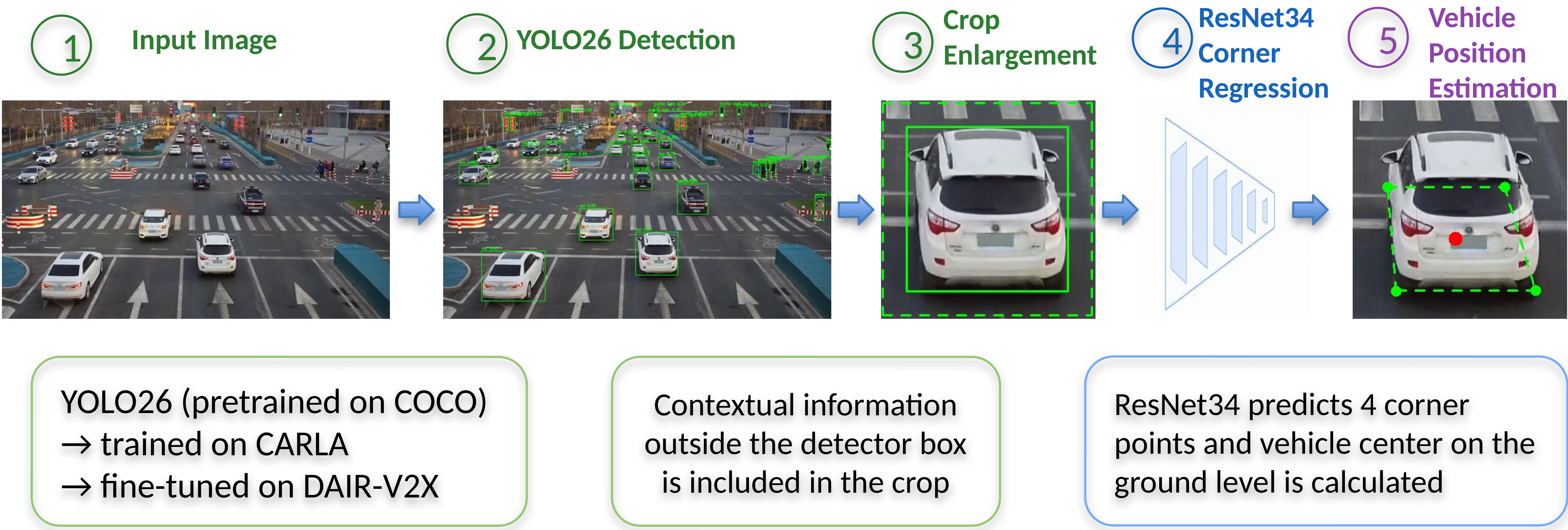}
    \caption{
    Overview of the proposed two-stage vehicle localization pipeline.
    Vehicles are first detected using YOLO26. Enlarged image crops are
    subsequently processed by a dedicated ResNet34 regression network
    predicting the projection of the vehicle base onto the road plane.
    The final vehicle position is estimated as the geometric center of
    the predicted quadrilateral.
    }
    \label{fig:pipeline}
\end{figure*}

\subsection{Overview}

The proposed approach consists of a two-stage processing pipeline designed to estimate the position of a vehicle on the road plane from a single monocular surveillance image. A schematic overview of the pipeline is presented in Fig.~\ref{fig:pipeline}.

In the first stage, vehicles are detected using a YOLO26-based object detector. For each detected object, the detector outputs a bounding box corresponding to the vehicle location in image space. These bounding boxes are subsequently used to generate image crops that serve as input to the second stage of the pipeline.

In the second stage, each cropped vehicle image is processed by a dedicated regression network that estimates the geometric structure of the vehicle projection on the road plane. More specifically, the network predicts four characteristic points corresponding to the projection of the vehicle base corners onto the image plane. These points define a quadrilateral approximating the contact region between the vehicle and the road surface.

The final vehicle position is estimated as the geometric center of the predicted quadrilateral. Compared with the standard bounding-box center representation, this approach provides a more accurate approximation of the true vehicle position on the road plane, particularly in the presence of strong perspective distortion and parallax effects.

The proposed architecture separates object detection and geometric regression into two independent stages. This design allows the regression model to focus exclusively on geometric estimation without being constrained by the objectives of the detection network. In addition, the decoupled formulation enables the use of enlarged image crops extending beyond the detector bounding box, providing additional contextual information useful for geometric reasoning.

\subsection{Vehicle Detection}

Vehicle detection is performed using the YOLO26 architecture \cite{yolo26_ultralytics}, which was selected due to its favorable trade-off between detection accuracy and computational efficiency. In the proposed pipeline, the lightweight YOLO26m variant was used in order to maintain real-time processing capability while preserving sufficient detection quality for subsequent geometric regression.

Although YOLO26 pretrained on the COCO dataset provides strong general-purpose detection performance, roadside surveillance imagery differs substantially from typical object detection benchmarks. In particular, roadside cameras introduce elevated viewpoints, strong perspective distortion, scale variation, truncation near image borders, and partial occlusions that are underrepresented in standard training datasets. As a consequence, direct application of a pretrained detector resulted in reduced localization quality for infrastructure-camera imagery.

To address this problem, the detector was additionally trained on synthetic CARLA data and subsequently fine-tuned on real-world DAIR-V2X imagery. The experimental results presented in Section~IV show that fine-tuning on real roadside imagery improves detection quality and matching IoU.

For each detected vehicle, the detector outputs an axis-aligned bounding box defined by image-space coordinates. These bounding boxes are used to extract image crops for the regression stage. To improve robustness against imperfect detections and to provide additional contextual information, an additional margin is applied around each bounding box before cropping.

\subsection{Corner Regression and Vehicle Position Estimation}

The second stage of the pipeline is formulated as a coordinate regression problem. A convolutional neural network based on the ResNet34 architecture \cite{He_2016_CVPR} is used to predict the image coordinates of four corner points corresponding to the projection of the vehicle base onto the road plane. Residual networks enable effective training of deep architectures through skip connections that mitigate the vanishing gradient problem.

ResNet34 was selected as a lightweight and computationally efficient backbone capable of extracting sufficiently rich geometric features while maintaining real-time applicability. The network input consists of cropped RGB vehicle images resized to a fixed spatial resolution.

The network output is represented as a vector of eight normalized values:

\begin{equation}
(x_1, y_1, x_2, y_2, x_3, y_3, x_4, y_4)
\end{equation}

where $(x_i, y_i)$ denote the image coordinates of the $i$-th corner point defining the projection of the vehicle base onto the road plane.

The regression targets are normalized relative to the cropped image dimensions. During inference, the predicted coordinates are transformed back into the original image coordinate system.

The predicted corner points define a quadrilateral approximating the contact region between the vehicle and the road surface. The final vehicle position is estimated as the geometric center of the predicted quadrilateral:

\begin{equation}
(x_c, y_c) =
\frac{1}{4}
\sum_{i=1}^{4}
(x_i, y_i)
\end{equation}

Unlike the center of the detector bounding box, the estimated quadrilateral center more closely corresponds to the true ground-plane location of the vehicle. This substantially reduces localization errors caused by perspective distortion and parallax, especially for large vehicles and elevated viewing angles.

The network is trained using the Mean Squared Error (MSE) loss applied to the predicted corner coordinates.

\subsection{Training Data Generation}

Synthetic training data were generated using the CARLA simulation environment \cite{dosovitskiy2017carla}. Eight different urban maps were used in order to increase scene diversity and improve generalization. Cameras were positioned at elevated roadside viewpoints resembling real surveillance-camera installations. All images were rendered at a resolution of $3840 \times 2160$ pixels.

All vehicle categories were merged into a single detection class. The object detector was trained on a synthetic dataset comprising 4748 images and 13\,577 annotated vehicle instances, divided into training, validation, and test subsets of 3345/730/673 images. The corner regression network was trained on a larger set of 59\,211 cropped vehicle instances (train/val/test: 51\,636/5\,570/2\,005), generated by applying data augmentation to the same source images.

Ground-truth corner annotations were generated automatically using the simulator-provided 3D bounding boxes together with camera intrinsic and extrinsic parameters. For each vehicle instance, the bottom face of the 3D bounding box was projected onto the image plane, yielding four corner points corresponding to the projection of the vehicle base onto the road surface.

\subsection{Real-World Data (DAIR-V2X)}

To evaluate the proposed method on real-world imagery, the infrastructure subset of the DAIR-V2X dataset \cite{dair} was used. DAIR-V2X is a large-scale dataset designed for vehicle--infrastructure cooperative perception and provides roadside camera images together with 3D object annotations and camera calibration parameters.

Four vehicle categories were retained for evaluation: Car, Van, Truck, and Bus. Ground-truth corner annotations were generated automatically from the provided 3D annotations. For each object instance, the 3D vehicle dimensions, orientation, object position, and virtual-LiDAR-to-camera transformation were used to reconstruct the vehicle bottom face in 3D space and project it onto the image plane.

For detector fine-tuning, 500 images containing 7231 bounding-box annotations were used, split into training, validation, and test subsets of 349/68/83 images, respectively. The regression network was fine-tuned using 3000 cropped vehicle samples, split into train/val/test subsets of 2100/450/450 samples.

Final evaluation was performed on a separate held-out subset consisting of 500 images and 7579 ground-truth vehicle instances, with no overlap with the fine-tuning data. Of these, 7108 instances were successfully matched through the full pipeline, while 223 were missed by the detector and 244 were rejected due to insufficient IoU between the predicted and ground-truth bounding boxes.

\subsection{Training Protocol}

Both stages of the pipeline were trained in two phases. In the first phase, the detector and regression network were trained using synthetic CARLA data. In the second phase, both models were fine-tuned on real-world DAIR-V2X imagery using the CARLA-trained weights as initialization.

This synthetic-to-real training strategy was adopted in order to exploit the large quantity of automatically generated synthetic annotations while adapting the models to the appearance characteristics and geometric properties of real roadside imagery. Experimental results indicate that fine-tuning on real-world data is essential for achieving high localization accuracy and compensating for the domain gap between synthetic and real imagery.

\subsection{Corner Regression Fine-Tuning}

\begin{figure}[h]
    \centering
    \includegraphics[width=\columnwidth]{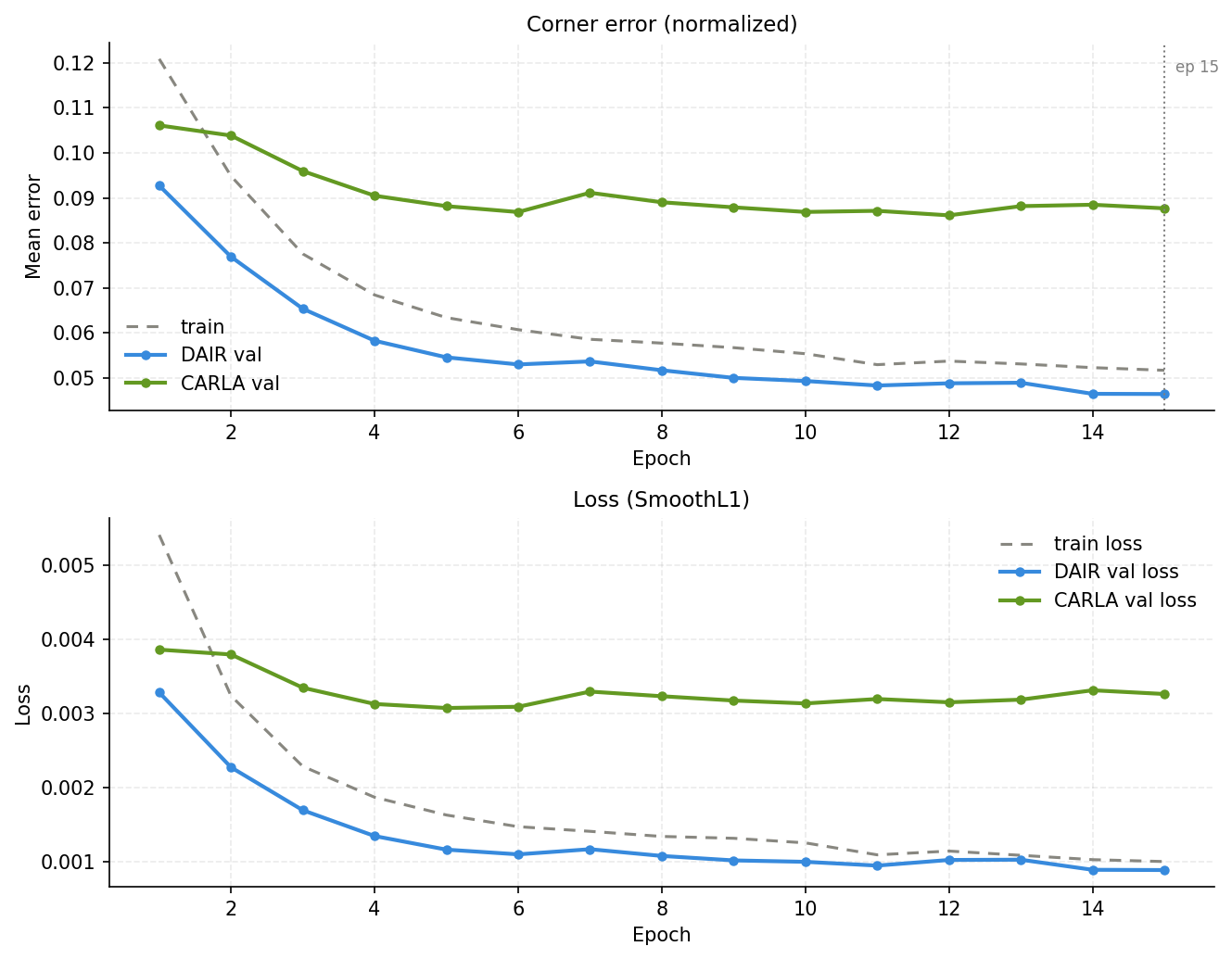} \caption{Corner regression fine-tuning curves on DAIR-V2X.  Both DAIR-V2X validation error (blue) and CARLA validation error (green) decrease consistently throughout training, confirming that fine-tuning on real-world data does not cause catastrophic forgetting of features learned on synthetic CARLA data.}
    \label{fig:finetune_history}
\end{figure}

To prevent catastrophic forgetting of CARLA-learned features  during fine-tuning, validation loss was monitored simultaneously on both the DAIR-V2X validation set and a held-out CARLA validation set after each epoch. Fig.~\ref{fig:finetune_history} shows the resulting training curves. Both validation curves decrease consistently throughout training without divergence, indicating that the DAIR-V2X annotations are geometrically consistent with the synthetic training data and that fine-tuning does not degrade generalization to previously learned viewpoints. The best checkpoint was selected based on DAIR-V2X validation loss.

\section{Experiments and Results}

\subsection{Evaluation Setup}

The proposed method was evaluated on both synthetic CARLA data and real-world DAIR-V2X imagery. The synthetic evaluation set contains 2005 vehicle instances generated in the CARLA simulation environment, while the real-world evaluation was performed on a held-out subset of the DAIR-V2X dataset containing 500 images and 7579 annotated vehicle instances.

The final evaluation includes only successfully matched detections, i.e., instances for which the detector produced a bounding box sufficiently overlapping the ground truth annotation. Matching was performed using Intersection over Union (IoU). Instances missed by the detector or rejected due to insufficient overlap were excluded from localization metrics. In total, 7108 vehicle instances were successfully matched through the complete pipeline, while 223 instances were missed by the detector and 244 were rejected due to insufficient IoU.

Two groups of evaluation metrics were used. First, pixel-domain localization error was computed in image space as the Euclidean distance between the predicted and ground-truth vehicle center projections. Second, for DAIR-V2X, ground-plane localization error was additionally evaluated in metric world coordinates using the provided camera calibration parameters. Match IoU mean denotes the average Intersection over Union computed exclusively over successfully matched detection--ground-truth pairs, providing a measure of bounding-box localization quality independent of recall.


To analyze localization robustness under varying visibility conditions, vehicle instances were divided into three visibility categories derived automatically from ground-truth annotations:
\begin{itemize}
    \item fully visible,
    \item occluded,
    \item truncated.
\end{itemize}

An object was classified as truncated if its bounding box intersected the image border within a 10-pixel tolerance margin. Occluded objects were identified using overlap between ground-truth bounding boxes. The resulting evaluation subset contained 2797 fully visible, 3701 occluded, and 610 truncated vehicle instances.

In addition, vehicle instances were grouped according to distance from the camera in order to analyze the influence of perspective distortion and parallax effects:
\begin{itemize}
    \item close range: $<40$ m,
    \item medium range: $40$--$80$ m,
    \item far range: $>80$ m.
\end{itemize}

\subsection{Detection Quality}
Table~\ref{tab:detection_quality} summarizes detector performance after successive training stages evaluated on the held-out DAIR-V2X subset (500 images, 7579 ground-truth instances). Three detector variants were evaluated:
\begin{itemize}
    \item YOLO26m pretrained on COCO,
    \item YOLO26m trained on synthetic CARLA data (4748 images, 13\,577 instances),
    \item YOLO26m fine-tuned on CARLA and DAIR-V2X (additional 500 images, 7231 instances).
\end{itemize}

The detector trained exclusively on synthetic CARLA imagery achieved lower matching IoU on real-world data than the original COCO-pretrained model, indicating a substantial synthetic-to-real domain gap despite the geometric realism of the simulator. Fine-tuning on real-world DAIR-V2X imagery significantly improved detection quality and yielded the highest matching IoU.

\begin{table}[h]
\centering
\caption{Detection quality on the held-out DAIR-V2X evaluation set (500 images, 7579 instances).}
\label{tab:detection_quality}
\begin{tabular}{lc}
\hline
Model & Match IoU mean \\
\hline
YOLO26m pretrained (COCO)        & 0.799 \\
YOLO26m trained (CARLA)          & 0.538 \\
YOLO26m fine-tuned (CARLA+DAIR)  & 0.888 \\
\hline
\end{tabular}
\end{table}

\subsection{Results on Synthetic Data (CARLA)}

On the CARLA evaluation set, the proposed method reduced the mean vehicle center localization error from 61.90~px to 15.56~px, corresponding to a 74.9\% improvement relative to the naive bounding-box-center baseline. The median error decreased from 54.05~px to 11.85~px (78.1\% reduction).

The mean IoU between the YOLO26m detector and ground-truth bounding boxes reached 0.953, indicating high-quality detections on synthetic CARLA imagery.

\begin{figure}[h]
    \centering
    \includegraphics[width=\columnwidth]{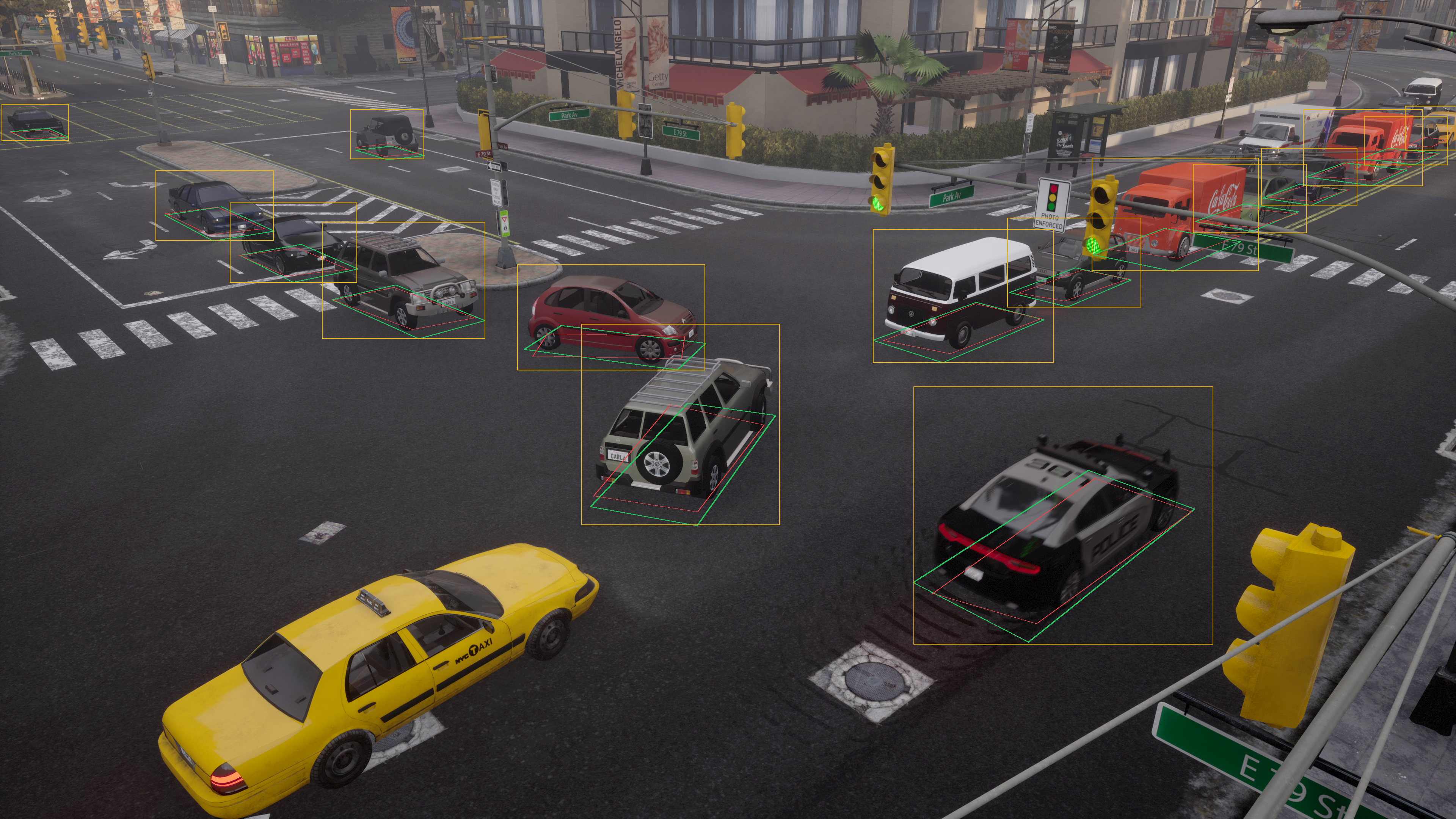}
    \caption{
    Example localization result on synthetic CARLA data.
    Green points indicate ground-truth corners, while red points indicate predicted corners.
    }
    \label{fig:carla_example}
\end{figure}

A representative localization result on synthetic CARLA data is shown in Fig.~\ref{fig:carla_example}.

\subsection{Results on Real Data (DAIR-V2X)}

On the DAIR-V2X evaluation subset, the proposed method reduced the mean image-space localization error from 31.77~px to 15.30~px, corresponding to a 51.8\% improvement over the naive bounding-box-center baseline. The median error decreased to 4.29~px, corresponding to a 79.7\% reduction relative to the naive median of 21.12~px.

After normalization by bounding-box height, the mean error decreased from 0.405 to 0.170, corresponding to a 58.0\% reduction. The normalized median error decreased from 0.340 to 0.066.

\begin{figure}[h]
    \centering
    \includegraphics[width=\columnwidth]{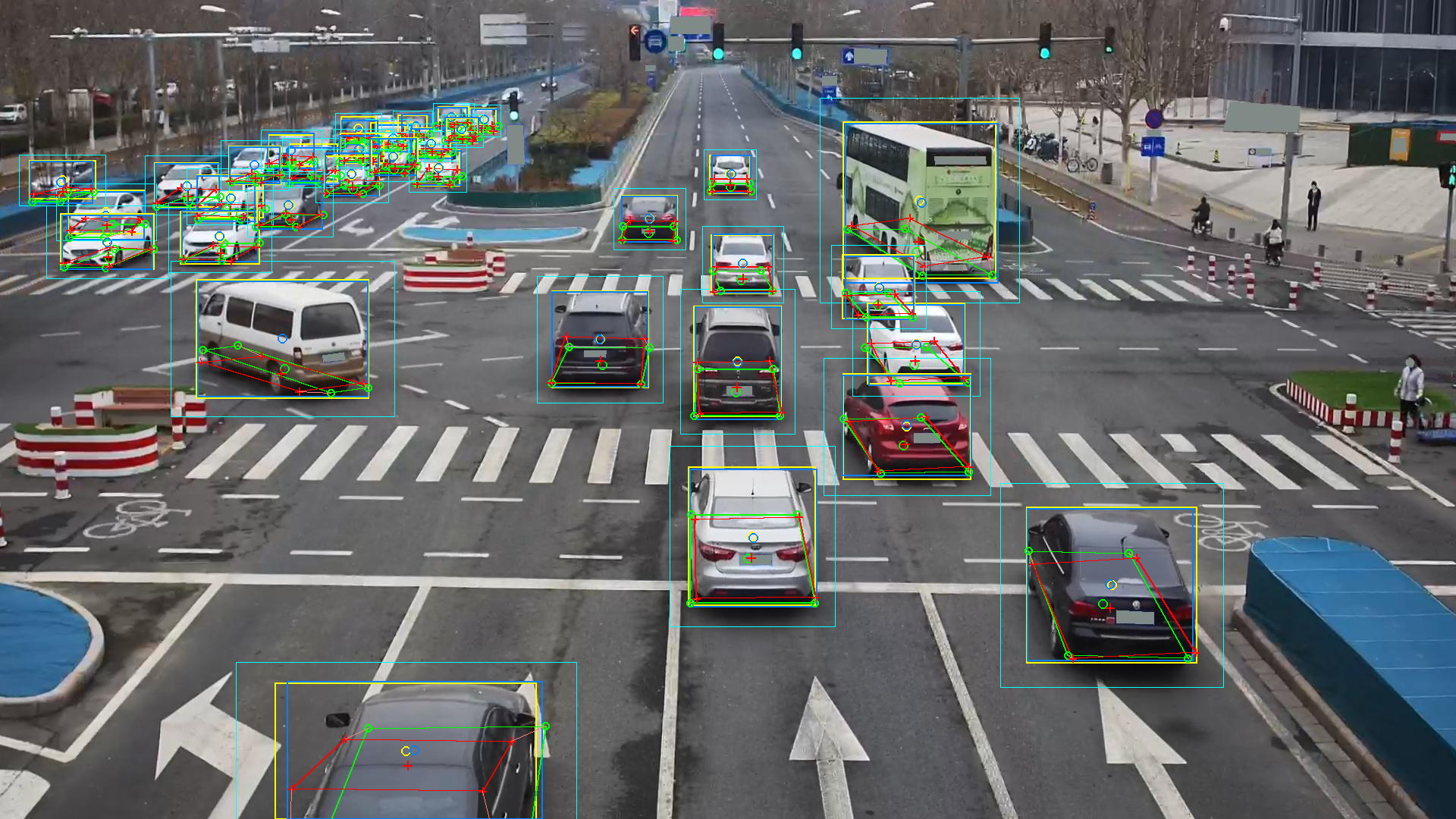}
    \caption{
    Example localization result on real-world DAIR-V2X imagery.
    Green points indicate ground-truth corners, while red points indicate predicted corners.
    }
    \label{fig:dair_example}
\end{figure}

\begin{figure}[h]
    \centering
    \includegraphics[width=\columnwidth]{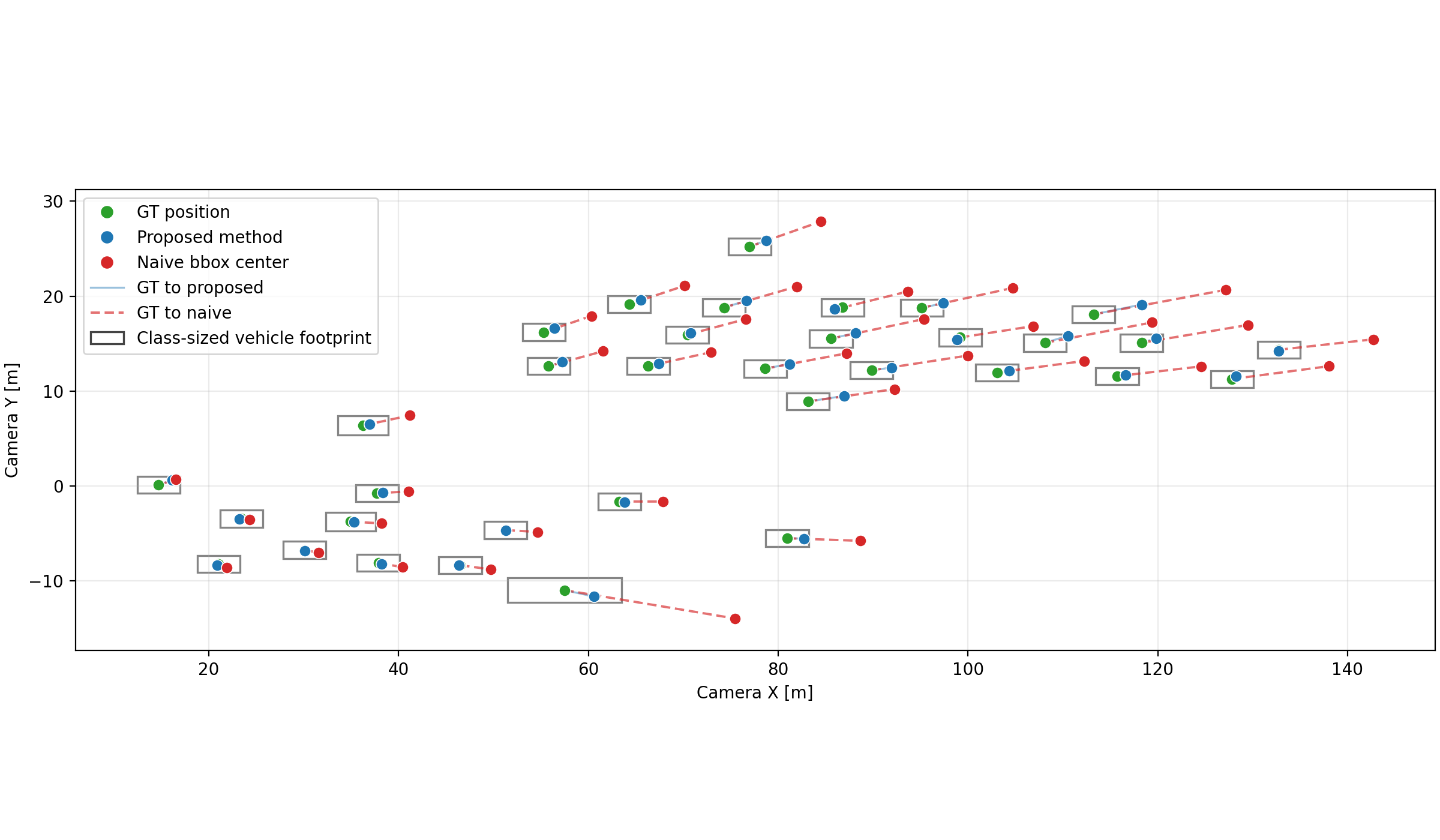}
    \caption{
    Bird’s-eye-view visualization of vehicle positions corresponding to Fig.~\ref{fig:dair_example}. Green points denote ground truth, blue points denote the proposed method, and red points denote the naive bounding-box-center baseline.
    }
    \label{fig:bev_example}
\end{figure}

Representative qualitative results are shown in Fig.~\ref{fig:dair_example}. Corresponding road-plane vehicle positions are visualized in Fig.~\ref{fig:bev_example}.

Table~\ref{tab:distance_error} reports ground-plane localization error as a function of vehicle distance from the camera. The proposed method consistently outperforms the naive baseline across all distance groups. The largest relative improvement is observed for medium- and far-range vehicles, where perspective distortion and parallax effects are strongest.

\begin{table}[h]
\centering
\caption{Ground-plane localization error by distance (DAIR-V2X).}
\label{tab:distance_error}
\begin{tabular}{lccccc}
\hline
Group & N & \multicolumn{2}{c}{Naive [m]} & \multicolumn{2}{c}{Proposed [m]} \\
 &  & mean & med. & mean & med. \\
\hline
Close ($<40$ m)      & 1139 & 2.42 & 2.04 & 1.26 & 0.91 \\
Medium ($40$--$80$ m) & 2877 & 5.95 & 5.52 & 1.14 & 0.90 \\
Far ($>80$ m)        & 3092 & 10.62 & 8.67 & 2.57 & 1.84 \\
\hline
\end{tabular}
\end{table}

Tables~\ref{tab:error_by_class} and~\ref{tab:error_by_visibility} summarize ground-plane localization error by vehicle type and visibility condition.

Tall vehicles, particularly trucks and buses, exhibit substantially higher localization errors due to stronger parallax effects and larger discrepancies between bounding-box geometry and true ground-plane projection. Nevertheless, the proposed method still significantly improves localization accuracy compared with the naive baseline, achieving over 75\% relative improvement across all vehicle categories.

Occluded vehicles achieve localization accuracy comparable to fully visible instances, indicating that the regression model generalizes well under partial occlusion. Truncated objects show lower relative improvement (44.6\%) compared with non-truncated instances (77.7\%), as parts of the projected vehicle footprint may lie outside the visible image region. This effect is partly caused by the spatial distribution of truncated objects within the dataset, as vehicles intersecting the image border are typically located close to the camera, where localization errors are generally smaller.

\begin{table}[h]
\centering
\caption{Ground-plane localization error by vehicle type (DAIR-V2X).}
\label{tab:error_by_class}
\begin{tabular}{lcccc}
\hline
Group & N & Naive [m] & \multicolumn{2}{c}{Proposed [m]} \\
 & & mean & mean & med. \\
\hline
Car              & 6209 &  6.22 & 1.54 & 1.05 \\
Van              &  451 &  8.94 & 2.27 & 1.57 \\
Truck            &  227 & 24.29 & 4.32 & 4.33 \\
Bus              &  221 & 20.54 & 5.02 & 3.88 \\
\hline
Low (Car+Van)    & 6660 &  6.40 & 1.59 & 1.08 \\
High (Truck+Bus) &  448 & 22.44 & 4.67 & 4.19 \\
\hline
\end{tabular}
\end{table}

\begin{table}[h]
\centering
\caption{Ground-plane localization error by visibility (DAIR-V2X).}
\label{tab:error_by_visibility}
\begin{tabular}{lcccc}
\hline
Group & N & Naive [m] & \multicolumn{2}{c}{Proposed [m]} \\
 & & mean & mean & med. \\
\hline
Fully visible     & 2797 & 7.88 & 1.96 & 1.17 \\
Occluded          & 3701 & 7.53 & 1.52 & 1.01 \\
Truncated         &  610 & 4.57 & 2.53 & 1.88 \\
All non-truncated & 6498 & 7.68 & 1.71 & 1.08 \\
\hline
\end{tabular}
\end{table}

\subsection{Ablation Study: Detection Margin}

To evaluate the influence of contextual information surrounding the detector bounding box, experiments were performed using different crop enlargement margins. The margin was applied symmetrically around the detected bounding box before generating the regression-network input crop.

\begin{figure}[h]
    \centering
    \includegraphics[width=\columnwidth]{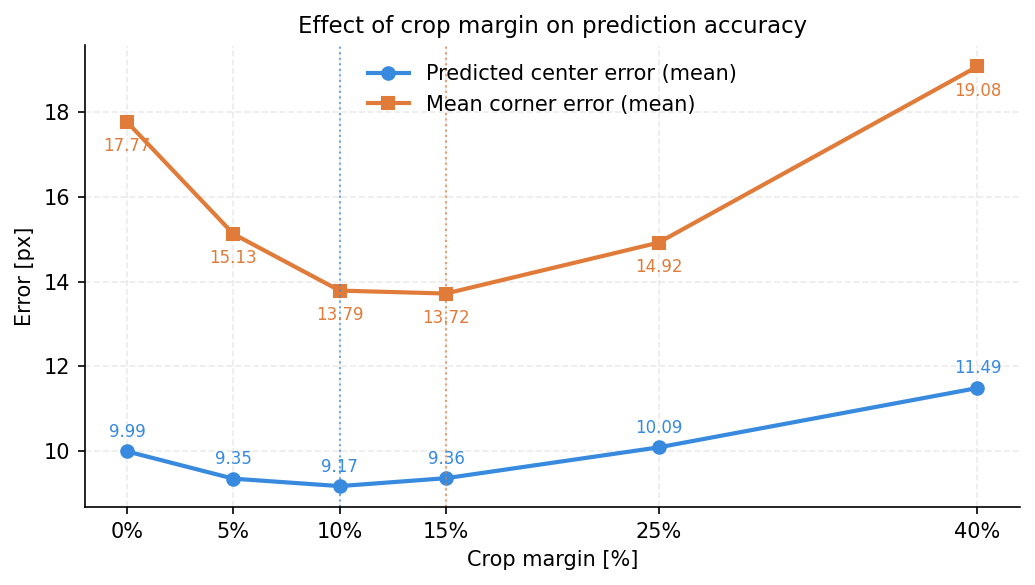}
    \caption{Effect of detection margin on predicted center error and mean corner error (DAIR-V2X). Optimal performance is achieved in the 10--15\% range.}
    \label{fig:margin_ablation}
\end{figure}

Table~\ref{tab:margin_ablation} and Fig.~\ref{fig:margin_ablation} summarize the obtained results. The best predicted center accuracy was achieved at a 10\% margin, while the lowest mean corner error was obtained at 15\%. Both metrics remain similar in the 10--15\% range, and a 15\% margin was selected for all subsequent experiments as it provides the best corner regression quality with only a marginal difference in center error.

\begin{table}[h]
\centering
\caption{Effect of detection margin on regression quality (DAIR-V2X).}
\label{tab:margin_ablation}
\begin{tabular}{lcc}
\hline
Margin & Predicted center [px] & Corner mean [px] \\
\hline
0\%  &  9.99 & 17.77 \\
5\%  &  9.35 & 15.13 \\
10\% &  \textbf{9.17} & 13.79 \\
15\% &  9.36 & \textbf{13.72} \\
25\% & 10.09 & 14.92 \\
40\% & 11.49 & 19.08 \\
\hline
\end{tabular}
\end{table}

The results indicate that contextual information outside the detector bounding box plays an important role in geometric localization. In many cases, cues relevant for estimating the vehicle footprint projection extend beyond the tight detector crop. These cues include road contact regions, vehicle shadows, wheel visibility, surrounding road geometry, and perspective relations relative to nearby objects.

At the same time, excessively large margins reduce localization accuracy by introducing unnecessary background structures and reducing the effective spatial resolution of the vehicle within the cropped image. This effect becomes particularly visible for distant vehicles occupying only a small fraction of the crop area.

Importantly, the matching IoU remained constant across all evaluated settings, confirming that the margin influences only the regression stage and does not affect detector performance. This confirms that the observed accuracy changes originate primarily from the geometric estimation process rather than from variations in detection quality.

\subsection{Runtime}
Inference time was evaluated for the complete localization pipeline, including YOLO-based object detection, crop extraction, corner regression, and postprocessing. Image loading time was excluded from the benchmark. Runtime measurements were performed on an NVIDIA GeForce RTX 4070 GPU using 1000 images from the DAIR-V2X-I dataset with a resolution of 1920×1080 pixels. Object detection was performed using the YOLO26m model with an input resolution of 1280 pixels.

The proposed method achieved an average processing time of 31.62 ms per image, corresponding to approximately 31.63 FPS. The average YOLO detection time was 22.00 ms per image. Crop extraction and preprocessing required 0.43 ms per object, while the corner regression network required 1.09 ms per object. Postprocessing required 0.03 ms per object.

To improve computational efficiency, corner regression was performed using batched GPU inference for all detected objects within a frame. The results indicate that the proposed localization framework operates in real-time conditions on a consumer-grade GPU.

\subsection{Discussion}

The experimental results confirm that regressing the projected vehicle base substantially reduces localization error compared with standard bounding-box-based representations. The improvement is consistent across distance groups, visibility conditions, and vehicle categories.

The largest improvements are observed for distant vehicles and for tall vehicles, where parallax effects are strongest. These results suggest that explicitly modeling the vehicle contact region on the road plane provides a more geometrically meaningful representation than using the detector bounding-box center alone.

The experiments additionally demonstrate that contextual information surrounding the detector bounding box plays an important role in geometric regression. A moderate crop enlargement improves localization accuracy, while excessively large margins introduce irrelevant background information.

The primary limitation of the current approach is handling truncated vehicles intersecting the image border. In such cases, parts of the projected vehicle footprint lie outside the visible image region, making accurate regression considerably more difficult. Although truncated objects represent only 7.6\% of the evaluation subset, they contribute disproportionately to the overall localization error.
An additional limitation of the current approach is reduced generalization to camera viewpoints substantially different from those present in the fine-tuning dataset. Although the regression network was initially trained on synthetic CARLA data generated using diverse camera perspectives, fine-tuning on DAIR-V2X imagery caused the model to adapt strongly to the geometric characteristics of the target dataset. As a result, the proposed method achieves high localization accuracy for viewpoints similar to those observed during fine-tuning, while performance may degrade for substantially different camera configurations or viewing angles.

\begin{figure}[h]
    \centering
    \includegraphics[width=.9\columnwidth]{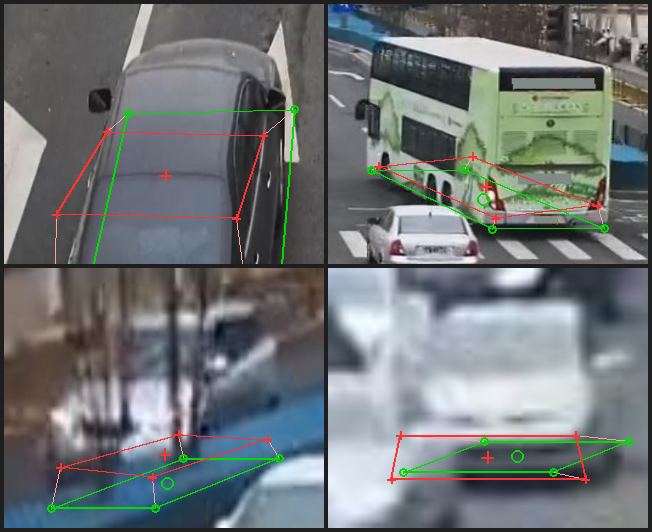}
    \caption{Representative failure cases. Top-left: truncated vehicle 
    intersecting image border — base corners extend outside the visible 
    region. Top-right: large bus affected by strong parallax — predicted 
    quadrilateral is geometrically distorted. Bottom-left: heavily 
    occluded vehicle — limited visibility degrades corner estimation. 
    Bottom-right: distant low-resolution vehicle — insufficient geometric 
    detail for accurate regression. Green indicates ground-truth corners, while red indicates predicted corners.}
    \label{fig:failure_cases}
\end{figure}

Representative challenging cases are shown in Fig.~\ref{fig:failure_cases}. Large vehicles such as buses constitute particularly demanding examples due to strong parallax effects and the relatively limited number of training samples available for these categories. In some cases, inaccuracies in corner regression cause the estimated vehicle position to be shifted further from the camera than the true ground-plane location. Nevertheless, even under such challenging conditions, the proposed method typically remains consistently more accurate than naive bounding-box-based localization approaches.

Distant low-resolution vehicles also remain challenging due to the limited amount of geometric information available in the image. Similarly, truncated vehicles intersecting the image border constitute a difficult failure mode because parts of the projected vehicle footprint may lie outside the visible image region. Partial occlusions generally have a smaller impact on localization accuracy, indicating that the regression model is able to exploit contextual and geometric cues even when parts of the vehicle are not directly visible.

\section{Conclusion}

This paper presented a geometry-aware monocular vehicle localization method for roadside surveillance cameras based on regression of the projected vehicle footprint. Unlike standard localization approaches relying directly on detector bounding-box geometry, the proposed method explicitly models the projection of the vehicle base onto the road plane using a dedicated corner regression network.

The proposed two-stage pipeline combines YOLO26-based object detection with ResNet34-based geometric regression operating on enlarged contextual image crops. By estimating the projected vehicle footprint rather than the detector center alone, the method significantly reduces localization errors caused by perspective distortion and parallax effects.

Experimental evaluation performed on both synthetic CARLA data and real-world DAIR-V2X imagery demonstrated significant improvements in localization accuracy compared with naive bounding-box-based approaches. The proposed method consistently improved both image-space and ground-plane localization accuracy across all evaluated distance ranges. 

The conducted experiments additionally showed that contextual information surrounding the detector bounding box plays an important role in geometric localization. Moderate crop enlargement improved regression accuracy, confirming that relevant geometric cues often extend beyond the tight detector crop.

The proposed method generalized well to partially occluded vehicles, indicating robustness to incomplete object visibility. The primary limitation of the current approach remains handling truncated vehicles intersecting image borders, where portions of the projected vehicle footprint are not visible. In some cases, large vehicles such as buses also remain challenging due to strong parallax effects and the relatively limited number of available training examples.

Overall, the results demonstrate that explicitly modeling vehicle footprint geometry provides an effective alternative to standard bounding-box-based localization methods.

Future work will focus on improving robustness for truncated and distant vehicles, extending the method to more diverse roadside camera configurations, and incorporating temporal information from video sequences to improve localization stability and tracking consistency.



\bibliographystyle{IEEEtran}
\bibliography{references}

\end{document}